\documentclass[10pt,twocolumn,letterpaper]{article}

\usepackage[pagenumbers]{cvpr} 

\usepackage{times}
\usepackage{epsfig}
\usepackage{graphicx}
\usepackage{amsmath}
\usepackage{amssymb}
\usepackage[font=small,labelfont=bf]{caption}
\usepackage{comment}
\usepackage{stfloats}
\usepackage{multirow}
\usepackage{float}
\usepackage{afterpage}
\usepackage[dvipsnames]{xcolor}
\usepackage[pagebackref,breaklinks,colorlinks]{hyperref}
\usepackage{silence}
\usepackage[capitalize]{cleveref}
\crefname{section}{Sec.}{Secs.}
\Crefname{section}{Section}{Sections}
\Crefname{table}{Table}{Tables}
\crefname{table}{Tab.}{Tabs.}

\begin{document}

\title{Forecasting Characteristic 3D Poses of Human Actions}

\author{
Christian Diller$^{1}$ \qquad\qquad Thomas Funkhouser$^{2}$  \qquad\qquad Angela Dai$^{1}$
\vspace{0.2cm} \\ 
$^{1}$Technical University of Munich \qquad $^{2}$Google 
}

\twocolumn[{%
	\renewcommand\twocolumn[1][]{#1}%
	\maketitle
	\begin{center}
		\vspace{-0.4cm}
		\captionsetup{type=figure}
		\includegraphics[width=\linewidth]{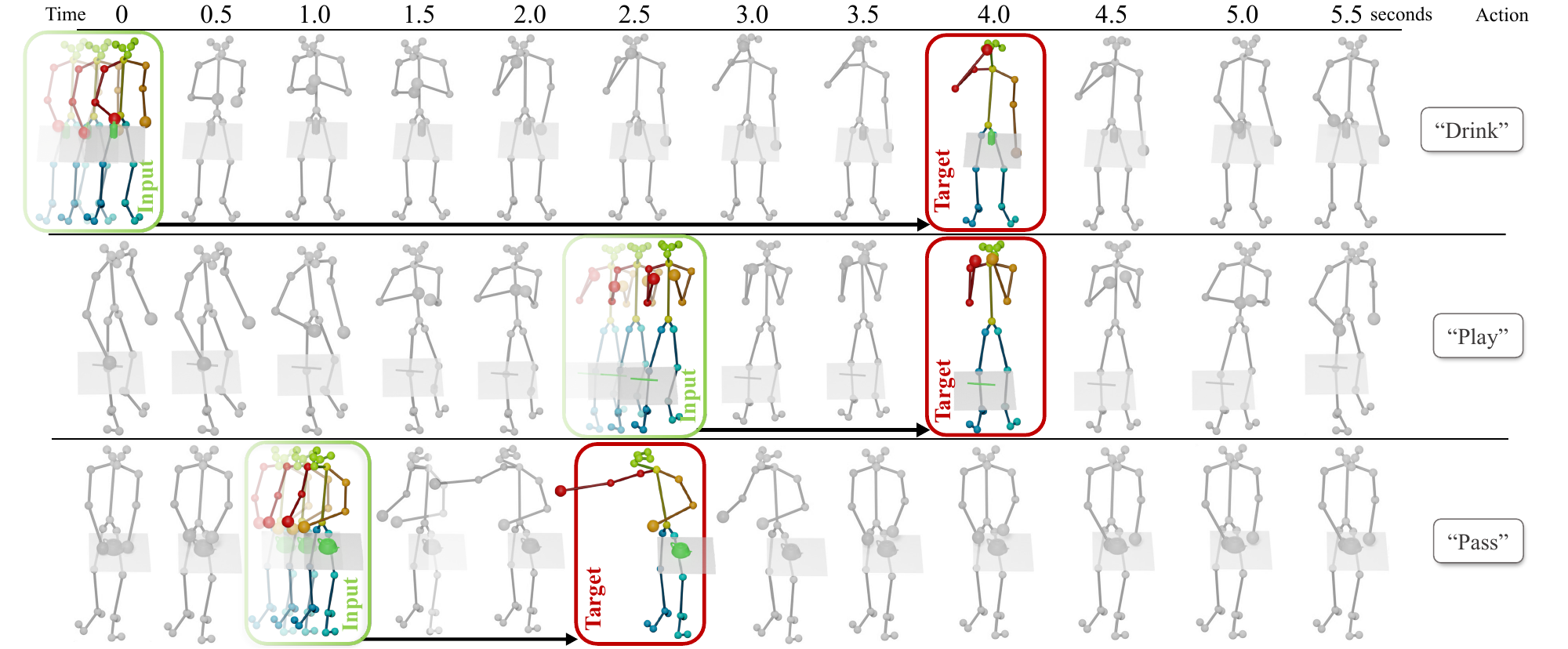}
		\captionof{figure}{
		For a real-world 3d skeleton sequence of a human performing an action, we propose to forecast the semantically meaningful \emph{characteristic 3d pose}, representing the {\color{Maroon} action goal} for this sequence. As input, we take a {\color{OliveGreen} short observation} of a sequence of consecutive poses leading up to the target characteristic pose. Thus, we propose to take a \emph{goal-oriented} approach, predicting the key moments characterizing future behavior, instead of predicting continuous motion, which can occur at varying speeds with predictions more easily diverging for longer-term ($>$1s) predictions.
		We develop an attention-driven probabilistic approach to capture the most likely modes of possible future characteristic poses.
		}
		\label{fig:teaser}
	\end{center}
}]
\maketitle

\begin{abstract}
\vspace{-0.25cm}
We propose the task of forecasting characteristic 3d poses: from a short sequence observation of a person, predict a future 3d pose of that person in a likely action-defining, characteristic pose -- for instance, from observing a person picking up an apple, predict the pose of the person eating the apple.
Prior work on human motion prediction estimates future poses at fixed time intervals.  Although easy to define, this frame-by-frame formulation confounds temporal and intentional aspects of human action.  Instead, we define a semantically meaningful pose prediction task that decouples the predicted pose from time, taking inspiration from  goal-directed behavior.
To predict characteristic poses, we propose a probabilistic approach that models the possible multi-modality in the distribution of likely characteristic poses.
We then sample future pose hypotheses from the predicted distribution in an autoregressive fashion to model dependencies between joints.
To evaluate our method, we construct a dataset of manually annotated characteristic 3d poses.
Our experiments with this dataset suggest that our proposed probabilistic approach outperforms state-of-the-art methods by 26\% on average.
\end{abstract}
\vspace{-0.25cm}

\section{Introduction}
\label{sec:intro}

\vspace{-0.1cm}
Future human pose forecasting is fundamental towards a comprehensive understanding of human behavior, and consequently towards achieving higher-level perception in machine interactions with humans, such as autonomous robots or vehicles.
In fact, prediction is considered to play a foundational part in intelligence \cite{bar2009proactive,hohwy2013predictive,clark2013whatever}.
In particular, predicting the 3d pose of a human in the future lays a basis for both structural and semantic understanding of human behavior, and for an agent to take fine-grained anticipatory action towards the forecasted future. For example, a robotic surgical assistant should predict in advance where best to place a tool to assist the surgeon’s next action, what sensor viewpoints will be best to observe the surgeon’s next manipulation, and how to position itself to be out of the way at critical future moments.

Recently, we have seen notable progress in the task of future 3D human motion prediction -- from an initial observation of a person, forecasting the 3D behavior of that person up to $\approx 1$ second in the future \cite{DBLP:conf/iccv/FragkiadakiLFM15,DBLP:conf/cvpr/JainZSS16,DBLP:conf/cvpr/MartinezB017,DBLP:conf/iccv/MaoLSL19, DBLP:conf/eccv/MaoLS20}.
Various methods have been developed, leveraging RNNs \cite{DBLP:conf/iccv/FragkiadakiLFM15,DBLP:conf/cvpr/JainZSS16,DBLP:conf/cvpr/MartinezB017,DBLP:conf/eccv/GuiWLM18}, graph convolutional neural networks \cite{DBLP:conf/iccv/MaoLSL19,DBLP:conf/cvpr/LiCZZW020}, and attention  \cite{DBLP:conf/ijcai/TangMLZ18,DBLP:conf/eccv/MaoLS20}.
However, these approaches all take a temporal approach towards forecasting future 3D human poses, and predict poses at fixed time intervals to imitate the fixed frame rate of camera capture.
This makes it difficult to predict longer-term (several seconds) behavior, which requires predicting both the time-based speed of movement as well as the higher-level goal of the future action.

\begin{figure}[b]
    \centering
    \vspace{-0.5cm}
    \includegraphics[width=\linewidth]{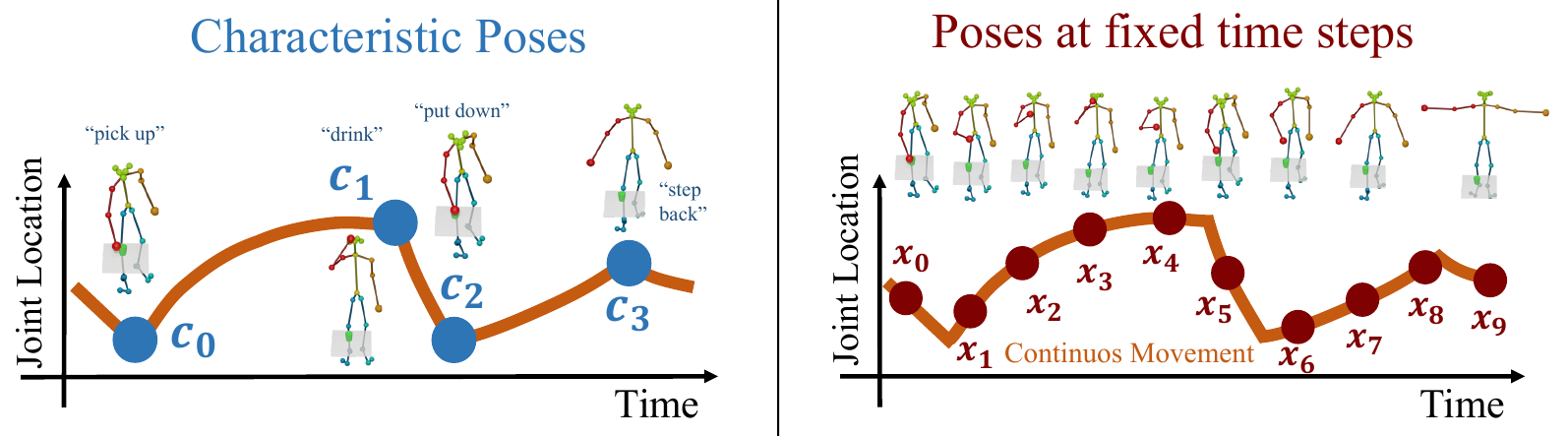}
    \vspace{-0.7cm}
    \caption{These plots show the salient difference between our new task (left) and the traditional one (right).   The orange curve depicts the motion of one joint (e.g., hand position as a person drinks from a glass).  It represents a typical piecewise continuous motion, which has discrete action-defining characteristic poses at cusps of the motion curves (e.g., grasping the glass on the table, putting it to ones mouth, etc.) separating smooth trajectories connecting them (e.g., raising or lowering the glass).  Our task is to predict future characteristic poses ({\color{MidnightBlue} blue dots} on left) rather than in-between poses at regular time intervals ({\color{Maroon} red points} on right).}
    \vspace{-0.25cm}
    \label{fig:charposes}
\end{figure}

Thus, we propose to decouple the temporal and intentional behavior, and introduce a new task of forecasting \emph{characteristic 3d poses} of a person's future action: from a short pose sequence observation of a human, the goal is to predict a future  pose of the person in a characteristic, action-defining moment.
This has many potential applications, including HRI, surveillance, visualization, simulation, and content creation. It could be used to predict the hand-off point when a robot is passing an object to a person; to detect and display future poses worthy of alerts in a safety monitoring system; to coordinate grasps when assisting a person lifting a heavy object; to assist tracking through occlusions; or to predict future keyframes, as is done in video generation \cite{DBLP:conf/iclr/JayaramanEEL19,pertsch2020keyframing}.

Fig.~\ref{fig:charposes} visualizes the difference between this new task and the traditional, time-based approach: our task is to predict a next characteristic pose at action-defining moments ({\color{MidnightBlue} blue dots}) rather than at fixed time-intervals ({\color{Maroon} red dots}).  As shown in Fig.~\ref{fig:teaser}, the characteristic 3d poses are more semantically meaningful and rarely occur at exactly the same times in the future.  
We believe that predicting possible future characteristic 3d poses takes an important step towards forecasting human action, by understanding the objectives underlying a future action or movement separately from the speed at which they occur.

Since future characteristic 3d poses often occur at longer-term intervals ($>1$s) in the future, there may be multiple likely modes of the characteristic poses, and we must capture this multi-modality in our forecasting.
Rather than deterministic forecasting, as is an approach in many 3D human pose forecasting approaches  \cite{DBLP:conf/iccv/MaoLSL19,DBLP:conf/eccv/MaoLS20,DBLP:conf/cvpr/LiCZZW020}, we develop an attention-driven prediction of probability heatmaps representing the likelihood of each human pose joint in its future location.
This enables generation of multiple, diverse hypotheses for the future pose. 
To generate a coherent pose prediction across all pose joints' potentially multi-modal futures, we make autoregressive predictions for the end effectors of the actions (e.g., predicting the right hand, then the left hand conditioned on the predicted right hand location) -- this enables a tractable modeling of the joint distribution of the human pose joints.

To demonstrate our proposed approach, we introduce a new benchmark on \emph{characteristic 3D Pose} prediction. 
We annotate characteristic keyframes in sequences from the GRAB~\cite{DBLP:conf/eccv/TaheriGBT20} and Human3.6M~\cite{DBLP:journals/pami/IonescuPOS14} datasets.
Experiments on this benchmark show that our probabilistic approach outperforms time-based state of the art by 26\% on average.

In summary, we present the following contributions:
\begin{itemize}
\itemsep0em
    \item We propose the task of forecasting \emph{characteristic 3D Poses}: predicting likely next action-defining future moments from a sequence observation of a person, towards goal-oriented understanding of pose forecasting.
	\item We introduce an attention-driven, probabilistic approach to tackle this problem and model the most likely modes for the next characteristic pose, and show that it outperforms state of the art.
	\item We autoregressively model the multi-modal distribution of future pose joint locations, casting pose prediction as a product of conditional distributions of end effector locations (e.g., hands), and the rest of the body.
	\item We introduce a dataset and benchmark on our \emph{characteristic 3D Pose} prediction, comprising 1535 annotated characteristic pose frames from the GRAB~\cite{DBLP:conf/eccv/TaheriGBT20} and Human3.6M~\cite{DBLP:journals/pami/IonescuPOS14} datasets.
\end{itemize} 

\section{Related Work}
\label{sec:related}

\begin{figure*}[t!]
\begin{center}
	\includegraphics[width=\linewidth]{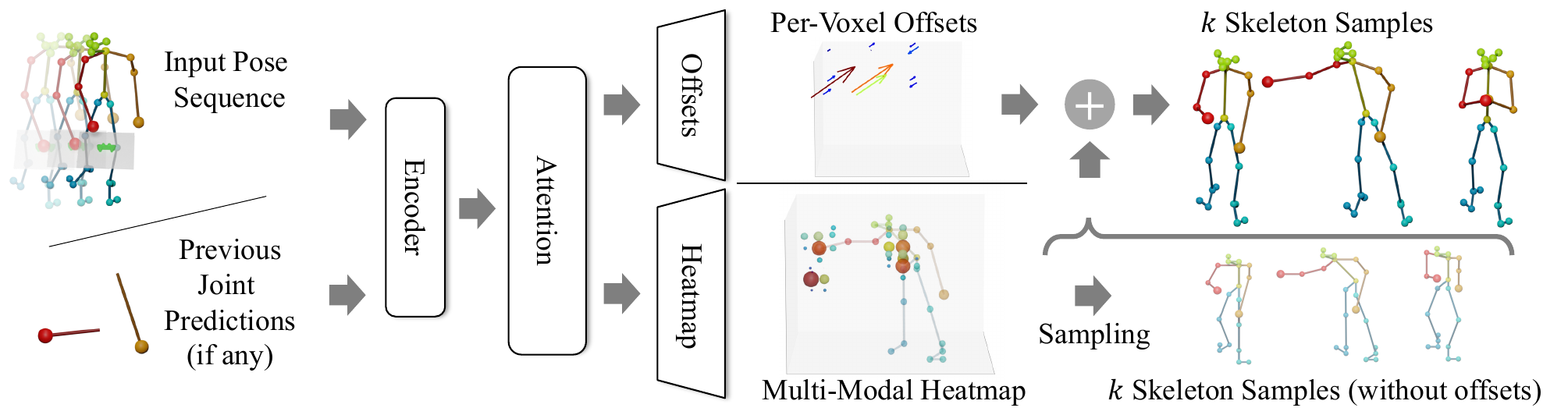}
\end{center}
\vspace{-0.6cm}
\caption{Overview of our approach for characteristic 3d pose prediction. 
From an input observed pose sequence, as well as any prior joint predictions, we leverage attention to learn inter-joint dependencies, and decode a 3d volumetric heatmap representing the probability distribution for the next joint to be predicted as well as a per-voxel offset field of same size for improved joint placement.
This enables autoregressive sampling to obtain final pose hypotheses characterizing likely characteristic 3d poses.}
\vspace{-0.25cm}
\label{fig:architecture}
\end{figure*}

\paragraph{Deterministic Human Motion Forecasting.}
Many works have focused on human motion forecasting, cast as a sequential task to predict a sequence of human poses according to the fixed frame rate capture of a camera.
For this sequential task, recurrent neural networks have been widely used for human motion forecasting \cite{DBLP:conf/iccv/FragkiadakiLFM15,DBLP:conf/cvpr/JainZSS16,DBLP:conf/cvpr/MartinezB017,DBLP:conf/iccv/AksanKH19,DBLP:conf/wacv/ChiuAWHN19,DBLP:conf/iccv/WangACHN19,DBLP:conf/cvpr/GopalakrishnanM19}.
Such approaches have achieved impressive success in shorter-term prediction (up to $\approx1$s, occasionally several seconds for longer term predictions), but the RNN summarization of history into a fixed-size representation struggles to maintain the long-term dependencies needed for forecasting further into the future.

To address some of the drawbacks of RNNs, non-recurrent models have also been adopted, encoding temporal history with convolutional or fully connected networks \cite{DBLP:conf/cvpr/ButepageBKK17,DBLP:conf/cvpr/LiZLL18,DBLP:conf/iccv/MaoLSL19}, or attention~\cite{DBLP:conf/ijcai/TangMLZ18,DBLP:conf/eccv/MaoLS20}.
Li et al.~\cite{DBLP:conf/iclr/ZhouLXHH018} proposed an auto-conditioned approach enabling synthesizing pose sequences up to $300$ seconds of periodic-like motions (walking, dancing).
However, these works all focus on frame-by-frame synthesis, with benchmark evaluation of up to $1000$ milliseconds.
Instead of a frame-by-frame synthesis, we propose a goal-directed task to capture perception of longer-term human action, which not only lends itself towards forecasting more semantically meaningful key moments, but enables a more predictable evaluation: as seen in Fig.~\ref{fig:teaser}, there can be significant ambiguity in the number of pose frames to predict towards a key or goal pose, making frame-based evaluation difficult in longer-term forecasting.

\vspace{-1em}
\paragraph{Multi-Modal Human Motion Forecasting.}
While 3d human motion forecasting has typically been addressed in a deterministic fashion, several recent works have introduced multi-modal future pose sequence  predictions. 
These approaches leverage well-studied approaches for multi-modal predictions, such as generative adversarial networks \cite{barsoum2018hp} and variational autoencoders \cite{yan2018mt,yuan2020dlow,aliakbarian2020stochastic}. 
For instance,  Aliakbarian et al.~\cite{aliakbarian2020stochastic} stochastically combines random noise with previous pose observations, leading to more diverse sequence predictions.
Yuan et al.~\cite{yuan2020dlow} learns a set of mapping functions which are then used for sampling from a trained VAE, leading to increased diversity in the sequence predictions than simple random sampling.
In contrast to these time-based approaches, we consider goal-oriented prediction of characteristic poses, and model multi-modality explicitly as predicted heatmaps for body joints in an autoregressive fashion to capture inter-joint dependencies. 

\vspace{-1em}
\paragraph{Goal-oriented Forecasting.}
While a time-based, frame-by-frame prediction is the predominant approach towards future forecasting tasks, several works have proposed to tackle goal-oriented forecasting.
Recently, Jayaraman et al.~\cite{DBLP:conf/iclr/JayaramanEEL19} proposed to predict ``predictable'' future video frames in a time-agnostic fashion, and represent the predictions as subgoals for a robotic tasks.
Pertsch et al.~\cite{pertsch2020keyframing} predict future keyframes representing a future video sequence of events.
Cao et al.~\cite{DBLP:conf/eccv/CaoGMCVM20} plan human trajectories from an image and 2d pose history, first predicting 2d goal locations for a person to walk to in order to synthesize the path.
Inspired by such goal-based abstractions, we aim to represent 3d human actions as its key, characteristic poses.
\vspace{-0.2cm}
\section{Method Overview}
\label{sec:overview}
\vspace{-0.2cm}

Given a sequence of $N$ 3d pose observations $\mathbf{X}_{1:N} = [\mathbf{x}_1, \mathbf{x}_2, ..., \mathbf{x}_N]$ of a person, our aim is to estimate a characteristic 3d pose of that person, characterizing the intent of the person's future action.
We take $J$ joint locations (represented as their 3d coordinates) for each pose of the input sequence, i.e. $\mathbf{x}_i\in \mathbb{R}^{J\times 3}$. From this input sequence, we predict a joint distribution of $J$ probability heatmaps $\mathbf{H}_j$ and finally, sample $K$ output pose hypotheses $\mathbf{Y}_{1:K}$, characterized by their $J$ 3d joints: $\mathbf{y}_i\in \mathbb{R}^{J\times 3}$. By representing probability heatmaps for the joint predictions, we can capture multiple different modes in likely characteristic poses, enabling more diverse future pose prediction. We note that we are the first to propose using volumetric heatmaps for future human pose forecasting, to the best of our knowledge, while previous work used them for the more deterministic task of pose estimation from multiple images \cite{iskakov2019learnable,tu2020voxelpose}.

From the input sequence, we develop a neural network architecture to predict a probability heatmap over a volumetric 3d grid for each joint, corresponding to likely future positions of that joint.
This enables effective modeling of multi-modality, but remains tied to a discrete grid, so we also regress a corresponding volume of per-voxel offsets, allowing for precise locations to be sampled.
Fig. \ref{fig:architecture} shows an overview of our learned probabilistic predictions.

We model these predictions conditionally in an autoregressive fashion in order to tractably model the joint distribution over all pose joint locations. 
This enables a consistent pose prediction over the set of pose joints, as a set of joints may have likely modes that are unlikely to be seen all together (e.g., right hand moving forward while the right elbow moves to the side -- both are valid independently but not together).
To sequentialize the pose joint prediction autoregressively, we first predict probability heatmaps for the end effectors in our dataset -- right hand first, then left hand conditioned on the right hand prediction, followed by the rest of the body joints. 

\section{Capturing Multi-Modality with Heatmap Predictions}
\label{sec:heatmap-prediction}

\begin{figure*}
\begin{center}
	\includegraphics[width=\linewidth]{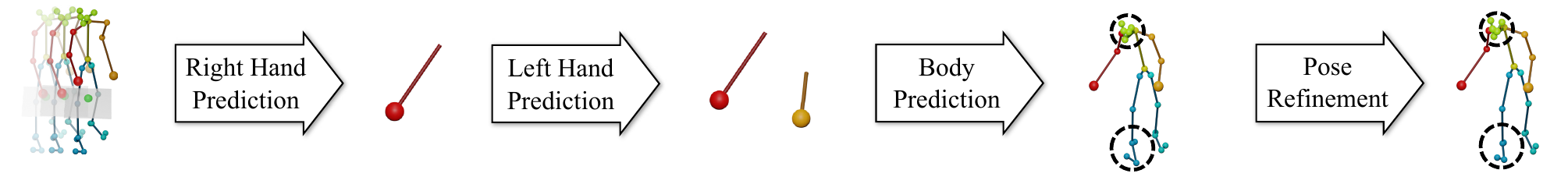}
\end{center}
\vspace{-0.8cm}
\caption{To model joint dependencies within the human skeleton, we sample joints in an autoregressive manner by first predicting the end-effectors (right and left hand), then the rest of the body; pose refinement then improves skeleton consistency.}
\vspace{-0.25cm}
\label{fig:autoregression}
\end{figure*}

We aim to learn to predict likely future locations for an output pose joint $j$, characterized by a probability heatmap  $\mathbf{H}_j$ over a volumetric grid of possible pose joint locations.
From the input sequence of $N$ pose observations of $J$ joints, and conditioned on any already predicted joints, we construct an attention-driven neural network to learn the different dependencies between human skeleton joints to inform the final heatmap prediction. 

\vspace{-1em}
\paragraph{Attention-Driven Sequence Encoding.}
We represent the body joints of the input sequence $\mathbf{X}_{1:N} = [\mathbf{x}_1, \mathbf{x}_2, ..., \mathbf{x}_N]$ as an $N\times J \times 3$ ($N=10$ as well as $J=25$ for the GRAB dataset and $J=17$ for Human 3.6M, respectively) concatenation of the joint locations over time. 
Features are first extracted with a single-layer GRU \cite{cho2014learning}.
We then compute an attention map from these features, representing dependencies to the input set of pose joints. 
This way, the network learns not only how different joints in the skeleton affect each other directly (e.g., kinematic relationships) but also learns to exploit more subtle correlations such as likely positions of one hand with respect to the other.
Following the formalism of Scaled Dot-Product Attention \cite{DBLP:conf/nips/VaswaniSPUJGKP17}, popularized in natural language processing, our attention maps are computed from a query $\boldsymbol{Q}$ and a set of key-value pairs $\boldsymbol{K}$ and $\boldsymbol{V}$. During training, representations for $\boldsymbol{Q}$, $\boldsymbol{K}$, and $\boldsymbol{V}$ are learned which are shared between all joints.  This allows us to project all joints into the same embedding space where we can then compare the joint of interest (represented by $\boldsymbol{Q}$) with all other joints ($\boldsymbol{K}$) to inform which parts of $\boldsymbol{V}$ (the learned latent representation for all joints which will be passed to the decoder) are relevant for this joint of interest.
\begin{equation}
\label{eq:attention}
\text{Attn}(\boldsymbol{Q}, \boldsymbol{K}, \boldsymbol{V}) = \text{softmax}\left(\frac{\boldsymbol{Q} \boldsymbol{K}^T}{\sqrt{D}}\right) \boldsymbol{V} = \boldsymbol{A} \boldsymbol{V}, 
\end{equation}
Intuitively, the similarity between key and query defines which parts of a learned pose skeleton representation are important for the desired prediction.
Formally, this is defined in Eq.~\ref{eq:attention}: The value representation $\boldsymbol{V}$ is weighed per-element by the result of the dot-product between $\boldsymbol{Q}$ and $\boldsymbol{K}$ (scaled by the dimension of the embedding vector $D$ and a softmax operation).
In our case, the attention map $\boldsymbol{A}$ has a dimensionality of $J'\times N$ with $J'$ indicating the number of joints to be predicted.
Any prior joint predictions for autoregressive prediction are considered as an additional node to our attention map, giving the attention map dimension $J'\times (N+n_p)$ for $n_p$ prior joints.

\vspace{-1em}
\paragraph{Heatmap Prediction.}
Based on the attention scoring, we then use a series of nine 3D convolutions to decode an output probability heatmap $\mathbf{H}_j$ for each body joint $j$. 
The grids are centered at the skeleton's hip joint; we use a grid size of $16^3$ voxels, spanning $2m^3$. 
A value in the grid of $\mathbf{H}_j$ at location $\mathbf{H}_j(x,y,z)$ corresponds to a probability of joint $j$ being at location $(x,y,z)$ in the future characteristic pose.
Instead of directly regressing the probability values, we predict $\mathbf{H}_j(x,y,z)$ as a classification problem by discretizing the output values into $n_{discr} = 10$ bins in the $[0,1]$ space.
We then use a cross entropy loss with the discretized target heatmap to train our heatmap predictions.
In our experiments, we found that this classification formulation for $\mathbf{H}_j$ produced better results than an $\ell_2$ or $\ell_1$ regression loss, as it mitigated tending towards the average or median.

\vspace{-1em}
\paragraph{Offset Prediction.}
Since predicting joint locations in a discrete grid inherently leads to grid artifacts in sampled output poses, we additionally learn an offset field $\mathbf{O}_j$ over the same volumetric grid. Here, each voxel $\mathbf{O}_j(x,y,z)\in \mathbb{R}^3$ represents the shift to be added after sampling a joint from the heatmap at $\mathbf{H}_j(x,y,z)$. We predict these offsets similarly to the heatmap volume, with a series of nine 3D convolutions, and clamp each offset vector $\mathbf{O}_j(x,y,z)$ to move the joint at most one voxel length.
Output poses are then estimated by sampling the heatmap, followed by refinement using the corresponding predicted offset.

\subsection{Training Details}
Note that for real-world data captured of human movement, we do not have a full ground truth probability distribution for the future characteristic pose, but rather a set of paired observations of input pose to the target pose.
Thus, we generate target heatmap data from a single future observation in the training data by applying a Gaussian kernel (size $5$, $\sigma=2$) over the target joint location.
At test time, we apply softmax scaling to the predicted heatmaps with a temperature of $0.025$ and from there, sample our final joint locations.
We learn multi-modality by generalizing across train set observations which results in seeing multiple possibilities for similar inputs (e.g., right vs. forward pass), encouraging learned heatmaps to represent multiple modes. We show that our formulation can effectively model multi-modal heatmaps in Section~\ref{sec:results}.

We train our models on a single NVIDIA GeForce RTX 2080Ti.
We use an ADAM optimizer with a weight decay of $0.001$ and a linear warmup schedule for $1000$ steps; learning rate is then kept at $0.001$.
We use a batch size of $100$, as a larger batch size helps with training our attention mechanism. 
Our model trains for up to $8$ hours until convergence.
During training, we apply teacher forcing, i.e. pose joint predictions conditioned on prior joint predictions are trained using the ground truth locations of the prior joints.
For a detailed specification of our network architecture, please refer to the appendix.

\section{Autoregressive Joint Prediction}
\label{sec:autoregression}

Given a set of heatmaps for each pose joint location, the next step is to predict specific joint locations.  Since they are not independent of one another, we cannot simply sample joint locations from each heatmap independently.  Instead, we must model the interdependencies between pose joints.

To do this, we model the joint distribution of pose joints autoregressively, as visualized in Fig. \ref{fig:autoregression}: 
we first predict end effector joints, followed by other body joints.
For our experiments, we find that the right and left hands tend to have a large variability, so we first predict the right hand, then the left hand conditioned on the right hand location, followed by the rest of the body joints. Empirically, we found that the hands tended to define the body pose, while the order of the rest has little impact.
To sample from a joint heatmap, we use temperature scaling to concentrate the heatmap near its local maxima, followed by random sampling.

\vspace{-1em}
\paragraph{Pose Refinement.}
While our autoregressive pose joint prediction encourages a coherent pose prediction with respect to coarse global structure, pose joints may still be slightly offset from natural skeleton structures.
Thus, we employ a pose refinement optimization to encourage the predicted pose to follow inherent skeleton bone length and angle constraints while keeping all joints in areas of high probability and the end-effectors close to their original prediction, as formulated in the objective function:
\vspace{-0.2cm}
\begin{multline}
\begin{aligned}
\text{E}_R&(\mathbf{x}, \mathbf{e}, \mathbf{b}, \mathbf{x_0}, \theta, H) =\\
&w_e\|\mathbf{x}_e-\mathbf{e}\|_2 + w_b\|\text{bonelengths}(x)-\mathbf{b}\|_1\\
&+ w_a\|\text{angles}(x)-\theta\|_1 + w_c\|x-x_0\|_1 \\ 
&+ w_h{\scriptstyle\sum}_j (1 - H_j)
\end{aligned}
\label{eq:refinement}
\end{multline}
where $\mathbf{x}$ the raw predicted pose skeleton as a vector of $N$ 3D joint locations; $\mathbf{b}$ and $\theta$ the bone lengths and joint angles, respectively, of the initially observed pose skeleton; $x_0$ the joint locations of the last skeleton in the input sequence; $H_j$ the heatmap probability for each joint; $\mathbf{e}$ the sampled end effector locations; and $w_e, w_b, w_a, w_h, w_c$ weighting parameters (in all our experiments, we use $w_e=0.2, w_b=1.0, w_a=0.4, w_h=0.1, w_c=0.1$).
We then optimize for $\mathbf{x}$ under this objective to obtain our final pose prediction.

\section{Characteristic 3D Pose Dataset}
\begin{figure}[bp]
\begin{center}
	\includegraphics[width=\columnwidth]{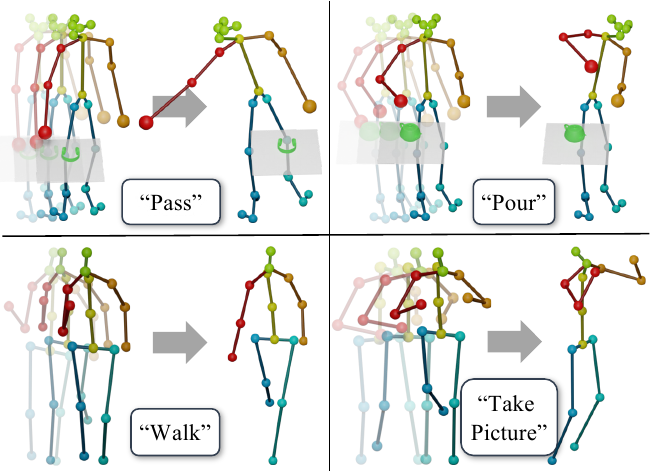}
\end{center}
\vspace{-0.6cm}
\caption{Example input observations and target characteristic 3d poses from our annotated datasets, based on GRAB (top) and Human3.6M (bottom).}
\vspace{-0.2cm}
\label{fig:dataset}
\end{figure}
To train and evaluate the task of characteristic 3d pose forecasting, we introduce a dataset of annotated characteristic poses, built on GRAB~\cite{DBLP:conf/eccv/TaheriGBT20} and Human3.6M~\cite{DBLP:journals/pami/IonescuPOS14}.
\begin{itemize}
    \item \textbf{Human3.6M} is a commonly used dataset for human pose forecasting, comprising 210 actions performed by 11 professional actors in 17 scenarios for a total of 3.6 million frames.
    3d locations are obtained for 32 joints via a high-speed motion capture system; we use a reduced 17-joint layout in our method, removing redundant and unused joints, following \cite{yuan2020dlow}.
    \item \textbf{GRAB} is a recent dataset with over 1 million frames in 1334 sequences of 10 different actors performing a total of 29 actions with various objects.
    Each actor starts in a T-Pose, moves towards a table with an object, performs an action with the object, and then steps back to the T-Pose.
    The human motions are captured using modern motion capture techniques, with an accuracy in the range of a few millimeters. GRAB provides SMPL-X \cite{DBLP:conf/cvpr/PavlakosCGBOTB19} parameters from which we extract the 25 most defining body joints. For more details, we refer to the appendix.
\end{itemize}

We then annotate the timesteps of the captured sequences corresponding to characteristic poses.
Input sequence start frames are randomly sampled, up until the characteristic pose frame.
Several example input sequence-characteristic pose pairs are visualized in Fig.~\ref{fig:dataset}.
Annotations were performed by the authors, within a time span of one day. This is the total time for annotating more that 1000 sequences across two datasets, with each annotation taking 10-30 seconds; this annotation efficiency enables quick and easy adoption of new datasets in the future.
We define a characteristic pose as the point in time when the action is most articulated, i.e. right before the actor starts returning back to another pose (e.g., when the hand is furthest from the person when passing, most tilted when pouring, etc.). For sequences containing multiple occurrences of the same action, like lifting, we chose the repetition with most articulation, e.g. when the object is lifted highest. In the case of Human3.6M, where there are sometimes multiple possible options for characteristic poses, we pick the first one that is representative of the action, e.g., the first sitting pose.

\vspace{-1em}
\paragraph{Characteristic 3D Pose Prediction.}
For the task of characteristic 3d pose prediction, we consider an input sequence of $N=10$ 3d pose observations of a person, represented as $J=25$ 3d joint locations for the GRAB dataset and $J=17$ for the Human3.6M dataset (in their native joint layouts; for more details we refer to the appendix).
From this observation, the next characteristic pose is predicted as $J$ 3d joint locations.
All poses are considered in their hip-centered coordinate systems. 
Note that while we have action labels in the annotated dataset, we do not use them for this task.

The $N$ input pose observations can occur at any time, so methods are trained with random input sequences up to the characteristic 3d pose.
At test time, five input points are evaluated for each method, with the five input points selected to evenly distribute between the beginning of the sequence to $N$ frames before the characteristic pose.

\vspace{-1em}
\paragraph{Evaluation.}
We use a train/val/test split by actor in each dataset.
For GRAB we have 8/1/1 train/val/test actors, resulting in 992/197/136 train/val/test sequences.
For Human3.6M, we follow the split of \cite{DBLP:conf/eccv/MaoLS20}: 5/1/1 and 150/30/30 train/val/test actors and sequences, respectively.

To evaluate our task of characteristic 3d pose prediction, we aim to consider the multi-modal nature of the task. Since we do not have ground truth probability distributions available, and only a single observed characteristic pose for each input pose observation, we follow previous work on multi-modal human pose sequence predictions \cite{yan2018mt,barsoum2018hp,yuan2020dlow,aliakbarian2020stochastic}: At test time, we consider $k=10$ hypotheses from each method.
To characterize these hypotheses holistically, we consider several metrics to assess accuracy, diversity, and quality of predictions.

\emph{Accuracy.} First, we evaluate the sampling error using the mean per-joint position error (MPJPE)~\cite{DBLP:journals/pami/IonescuPOS14} by comparing the most similar prediction $p'$ to the ground-truth pose $p$:
\begin{equation}
\vspace{-0.3cm}
\label{eq:mpjpe}
\text{E}_{\text{MPJPE}} = \frac{1}{N}\sum_{j=1}^{N} ||p'_j - p_j||_2^2
\end{equation}
This evaluates whether the predicted hypotheses capture the target well and allows for comparison with deterministic baselines (where all hypotheses are identical). 

\emph{Diversity.} We evaluate the diversity as the MPJPE between all sampled poses for the same sequence. This evaluates the multi-modality of predicted distributions.

\emph{Quality.} Finally, we evaluate quality of our multi-modal predictions with the Inception Score \cite{salimans2016improved} (IS) over the set of predicted hypotheses for all test sequences. The Inception Score is widely used to measure the quality generative model outputs. More specifically, we use the conditional formulation first introduced in \cite{huang2018multimodal}.
Similar to \cite{aliakbarian2020stochastic}, we adapt this idea to our use case by training a simple skeleton-based action classifier on ground-truth samples from our datasets. Overall, this metric estimates how well the predictions capture an action while still producing diverse poses.

\section{Experimental Evaluation}
\label{sec:results}

\begin{table*}[tp]
\begin{center}
	\begin{tabular}{| c | l || c | c | c || c | c | c |}
		\hline
		\multicolumn{2}{|c||}{} & \multicolumn{3}{c||}{\footnotesize GRAB} & \multicolumn{3}{c|}{\footnotesize Human3.6m} \\ \hline
		& Method & \footnotesize MPJPE $\downarrow$ & \footnotesize Diversity $\uparrow$ & \footnotesize IS $\uparrow$ & \footnotesize MPJPE $\downarrow$ & \footnotesize Diversity $\uparrow$ & \footnotesize IS $\uparrow$ \\ \hline\hline
		\multirow{3}{*}{\rotatebox[origin=c]{90}{\footnotesize Statistical}} & Random Sampling & 1.018 & - & - & 1.159 & - & - \\ \cline{2-8}
		& Average Train Pose & 0.146 & - & - & 0.179 & - & - \\ \cline{2-8}
		& Zero Velocity & 0.063 & - & - & 0.166 & - & - \\ \cline{2-8}\hline\hline
        \multirow{4}{*}{\rotatebox[origin=c]{90}{\footnotesize Algorithmic}} & Learning Trajectory Dependencies~\cite{DBLP:conf/iccv/MaoLSL19} & 0.077 & - & - & 0.165 & - & - \\ \cline{2-8}
        & History Repeats Itself~\cite{DBLP:conf/eccv/MaoLS20} & 0.071 & - & - & 0.116 & - & - \\ \cline{2-8}
        & DLow~\cite{yuan2020dlow} & 0.071 & 0.089 & 1.257 {\footnotesize $\pm 0.02$} & 0.119 & 0.104 & 1.623 {\footnotesize $\pm 0.08$} \\ \cline{2-8}
        & \textbf{Ours} & \textbf{0.054} & {\bf 0.105} & {\bf 4.153 {\footnotesize $\pm 0.87$}} & \textbf{0.092} & {\bf 0.189} & {\bf 3.139 {\footnotesize $\pm 0.32$}} \\ \hline
	\end{tabular}
	\vspace{-0.2cm}
	\caption{Characteristic 3d pose performance, in comparison with state of the art and statistical baselines. We evaluate MPJPE for all methods and additionally, the diversity of multi-modal methods in terms of MPJPE between samples as well as their quality with the Inception Score, similar to \cite{aliakbarian2020stochastic}.}
	\label{tab:comparison}		
\end{center}
\vspace{-0.7cm}
\end{table*}

\begin{table}[b]
\vspace{-0.5cm}
\begin{center}
    \resizebox{\columnwidth}{!}{%
	\begin{tabular}{| l || c | c || c | c |}
		\hline
		Method & \multicolumn{2}{c|}{\footnotesize GRAB} & \multicolumn{2}{c|}{\footnotesize Human3.6m} \\ \hline\hline
		 & \footnotesize MPJPE $\downarrow$ & \footnotesize IS $\uparrow$ & \footnotesize MPJPE $\downarrow$ & \footnotesize IS $\uparrow$ \\ \hline\hline
        L. T. D.~\cite{DBLP:conf/iccv/MaoLSL19} & 0.075 & - & 0.156 & - \\ \hline
        H. R. I.~\cite{DBLP:conf/eccv/MaoLS20} & 0.066 & - & 0.116 & - \\ \hline
        DLow~\cite{yuan2020dlow} & 0.059 & 1.567 {\footnotesize $\pm 0.02$} & 0.108 & 1.418 {\footnotesize $\pm 0.14$} \\ \hline\hline
        \textbf{Ours} & \textbf{0.054} & {\bf 4.153 {\footnotesize $\pm 0.87$}} & \textbf{0.092} & {\bf 3.139 {\footnotesize $\pm 0.32$}} \\ \hline
	\end{tabular}
	}
	\caption{Characteristic 3d pose performance comparison. In contrast to Tab \ref{tab:comparison}, baselines are provided with ground-truth characteristic time step information.}
	\label{tab:oracle-comparison}		
\end{center}
\vspace{-0.5cm}
\end{table}

We evaluate the task of characteristic 3d pose prediction, using our annotated dataset built from the real-world GRAB~\cite{DBLP:conf/eccv/TaheriGBT20} and Human3.6M~\cite{DBLP:journals/pami/IonescuPOS14} datasets.

\paragraph{Comparison to time-based state-of-the-art forecasting.}
In Tab.~\ref{tab:comparison}, we compare to state-of-the-art multi-modal sequence forecasting approach DLow \cite{yuan2020dlow}, which is based on a conditional VAE, as well as to recent deterministic approaches for frame-based future human motion prediction, Learning Trajectory Dependencies~\cite{DBLP:conf/iccv/MaoLSL19} and History Repeats Itself~\cite{DBLP:conf/eccv/MaoLS20}, which use a graph neural network and an attention-based model, respectively, to predict human pose sequences. 
We train all of these sequential approaches on our datasets, given the input sequence of $N$ frames, to predict an output $N_o$-frame pose sequence, with $N_o=100$ frames  to ensure that the characteristic pose falls within each target sequence.
Since these sequence-based approaches each predict output sequences, we additionally allow them to predict the time step of the characteristic pose with an MLP to obtain the final characteristic pose prediction (see the appendix for additional detail).

Since we aim to predict a characteristic 3d pose given an arbitrary sequence observation, we sample different start points for the input sequence, and analyze performance across varying distance from the goal pose.

\begin{figure}[b]
\vspace{-0.5cm}
\begin{center}
    \includegraphics[width=\columnwidth]{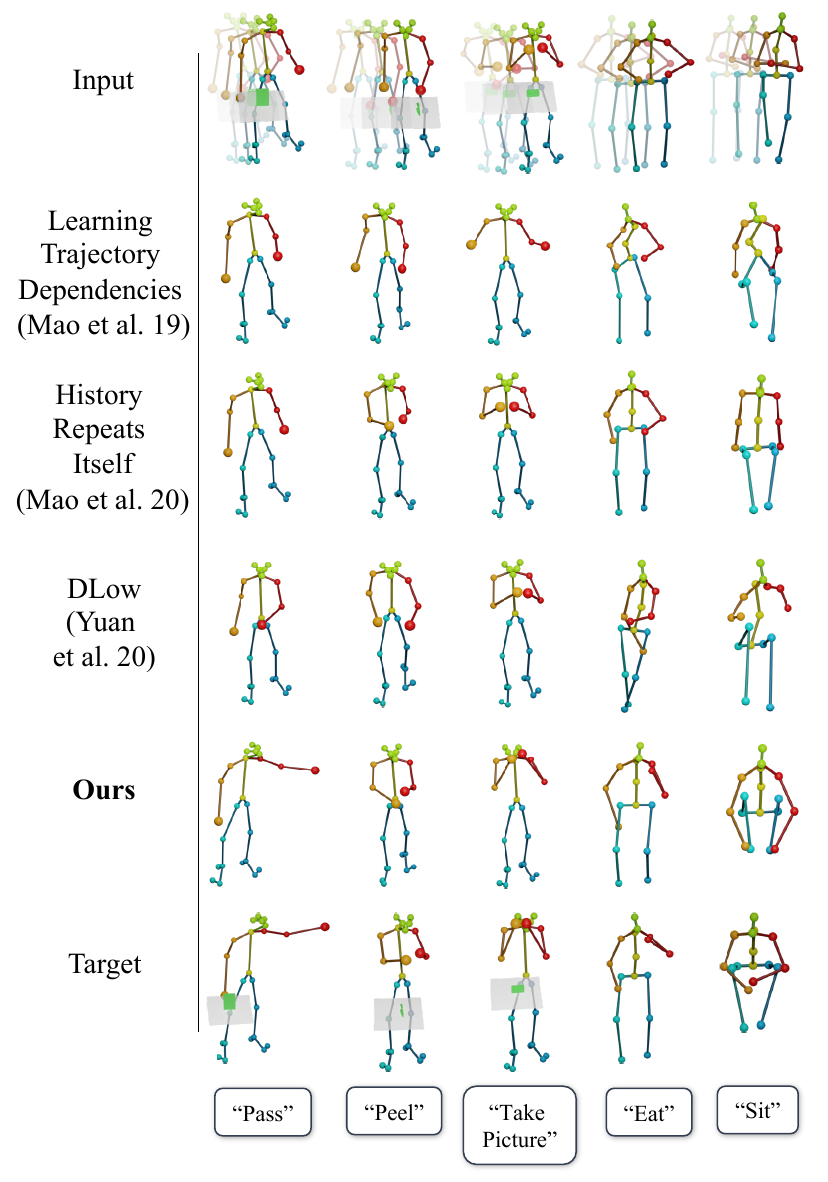}
\end{center}
\vspace{-0.7cm}
\caption{Qualitative results on characteristic 3d pose prediction. In comparison to deterministic  \cite{DBLP:conf/iccv/MaoLSL19,DBLP:conf/eccv/MaoLS20} (rows 2 and 3) and probabilistic \cite{yuan2020dlow} (row 4) approaches, our method more effectively predicts likely intended action poses. Note that action labels are only shown for visualization purposes.}
\label{fig:gallery-baselines}
\end{figure}

We report the MPJPE, Diversity, and IS metrics in Tab. \ref{tab:comparison}; we first measure the performance for each of the five input sequence start times mentioned above and average over those for the final result. Our approach more accurately characterizes the future characteristic poses while also producing improved diversity and quality.
For comparison, we also report baseline performance when given an oracle providing the ground-truth characteristic time step in Tab. \ref{tab:oracle-comparison}. 
Even with this additional information, our characteristic pose formulation achieves improved results.
Qualitative results are shown in Fig.~\ref{fig:gallery-baselines}; our probabilistic approach more effectively captures a realistic set of characteristic modes.

In Fig. \ref{fig:multi-modality}, we  visualize the diversity of our predictions in comparison with  multi-modal baselines. Our predicted pose hypotheses show more diversity  in both joint placement and action representation, while still capturing the target pose.

\vspace{-1em}
\paragraph{Comparison to statistical baselines.}
We also compare with three statistical baselines: full random sampling from an evenly distributed heatmap, the average target train pose over the entire dataset, and a zero-velocity baseline (i.e., the error of simply using the last input pose as prediction), which was shown by Martinez et al.~\cite{DBLP:conf/cvpr/MartinezB017} to be competitive with and sometimes outperform state of the art.
Our approach outperforms these statistical baselines, indicating learning of strong characteristic pose patterns.

\section{Ablation Studies}
\label{sec:ablations}

\paragraph{Does a probabilistic prediction help?}
In addition to comparing to state-of-the-art alternative approaches which make deterministic predictions, we compare in Tab.~\ref{tab:ablation} with our model backbone with a deterministic output head (an MLP) replacing the volumetric heatmap decoder which regresses offset positions for each pose joint relative to the input positions.
Removing our heatmap predictions similarly fails to effectively capture the characteristic modes; our probabilistic, heatmap-based predictions notably improve performance.

\begin{figure*}
\vspace{-0.75cm}
    \centering
    \includegraphics[width=\linewidth]{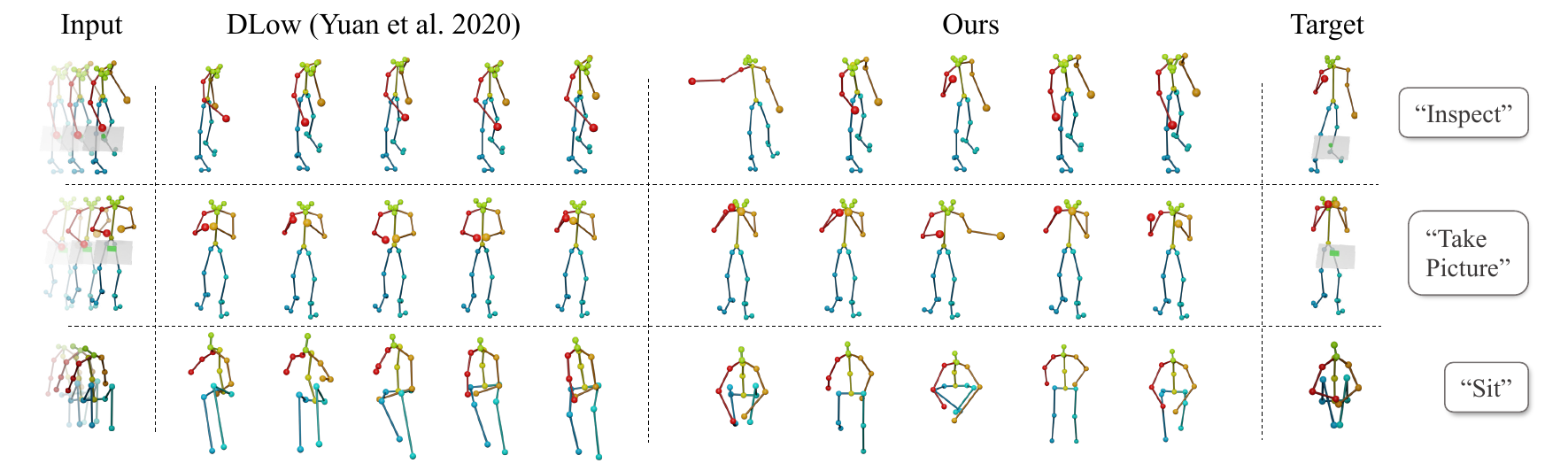}
    \vspace{-0.8cm}
    \caption{Qualitative results on characteristic 3d pose prediction, showing the diversity of our predictions in comparison with DLow \cite{yuan2020dlow}.}
    \label{fig:multi-modality}
    \vspace{-0.3cm}
\end{figure*}

\vspace{-1em}
\paragraph{Does per-voxel offset prediction help?}
We analyze the effect of per-voxel offset prediction in Tab.~\ref{tab:ablation}, showing that they notably improve pose predictions.
Applying pose refinement without offset prediction fails to achieve the same level of improvement.

\begin{table}[b]
\vspace{-0.25cm}
    \centering
    \resizebox{\columnwidth}{!}{%
	\begin{tabular}{| c | l || c | c || c | c |}
		\hline
		& & \multicolumn{2}{c|}{\footnotesize GRAB} & \multicolumn{2}{c|}{\footnotesize Human3.6m} \\ \hline\hline
		& Ablation & \footnotesize MPJPE $\downarrow$ & \footnotesize IS $\uparrow$ & \footnotesize MPJPE $\downarrow$ & \footnotesize IS $\uparrow$ \\ \hline\hline
		\multirow{2}{*}{\rotatebox[origin=c]{90}{\footnotesize Loss}} 
        & $\ell_1$ loss & 0.132 & 1.132 {\footnotesize $\pm 0.01$} & 0.198 & 2.246 {\footnotesize $\pm 0.24$} \\ \cline{2-6}
        & $\ell_2$ loss & 0.130 & 1.146 {\footnotesize $\pm 0.01$} & 0.206 & 1.976 {\footnotesize $\pm 0.08$} \\ \hline\hline
        \multirow{2}{*}{\rotatebox[origin=c]{90}{\footnotesize Model}} 
        & Deterministic & 0.064 & - & 0.108 & - \\ \cline{2-6}
        & Not autoreg. & 0.077 & 1.583 {\footnotesize $\pm 0.15$} & 0.109 & 1.929 {\footnotesize $\pm 0.09$} \\ \hline\hline
        \multirow{4}{*}{\rotatebox[origin=c]{90}{\footnotesize Sampling}} 
        & No offsets & 0.132 & 1.328 {\footnotesize $\pm 0.02$} & 0.172 & 2.537 {\footnotesize $\pm 0.07$} \\ \cline{2-6}
        & $\hookrightarrow$ refined \hfill & 0.127 & 1.509 {\footnotesize $\pm 0.03$} & 0.163 & 2.978 {\footnotesize $\pm 0.14$} \\ \cline{2-6}
        & $k=50$ & 0.049 & 1.222 {\footnotesize $\pm 0.02$} & 0.082 & 1.845 {\footnotesize $\pm 0.19$} \\ \cline{2-6}
        & Not refined & 0.057 & 3.989 {\footnotesize $\pm 0.95$} & 0.098 & 2.418 {\footnotesize $\pm 0.11$} \\ \hline\hline
        & \textbf{Ours} & \textbf{0.054} & {\bf 4.153 {\footnotesize $\pm 0.87$}} & \textbf{0.092} & {\bf 3.139 {\footnotesize $\pm 0.32$}} \\ \hline
	\end{tabular}
	}
	\vspace{-0.2cm}
	\caption{Ablation study over varying heatmap losses, deterministic and non-autoregressive pose sampling, no offset prediction (with and without pose refinement), number of samples taken for the evaluation, and without pose refinement.}
	\label{tab:ablation}
\end{table}

\vspace{-1em}
\paragraph{Does autoregressive pose joint sampling help?}
We analyze the effect of our autoregressive pose joint sampling in Tab.~\ref{tab:ablation}.
We compare against a version of our model trained to predict each pose joint heatmap independently, with pose joints sampled independently, which often results in valid individual pose joint predictions that are globally inconsistent with the other pose joints.
In contrast, our autoregressive sampling helps to generate a likely, consistent pose.

\vspace{-1em}
\paragraph{How diverse are the sampled poses?}
We show qualitative examples of our multi-modal predictions in Fig. \ref{fig:multi-modality}, outlining the diversity of both heatmap predictions and sampled skeletons. 
We also evaluate our prediction diversity as MPJPE between our sampled outputs as part of Tab. \ref{tab:comparison}.

\vspace{-1em}
\paragraph{What is the effect of the number of pose samples?}
If we take more pose samples from our predicted joint distribution (from $10$ to $50$), we can, as expected, better predict the potential target characteristic pose, as seen in Tab. \ref{tab:comparison}.

\vspace{-1em}
\paragraph{Do different heatmap losses matter?}
We evaluate our formulation for heatmap prediction as a discretized heatmap with a cross entropy loss against regressing heatmaps with an $\ell_1$ or $\ell_2$ loss, and find that our discretized formulation much more effectively models the relevant modes.

\vspace{-1em}
\paragraph{Limitations.}
Several limitations remain for our approach of characteristic 3d action pose forecasting. For instance, while our offset predictions help alleviate the ties to a volumetric heatmap grid, more precise modeling of smaller-scale behavior (e.g., detailed hand movement) would require more efficient representations such as sparse grids.
In addition, our method relies on manually annotated characteristic 3d poses for supervision; while characteristic pose annotation is very efficient for new datasets, self-supervised formulations would also be an interesting future direction.

\section{Conclusion}
\label{sec:conclusion}
In this paper, we introduced a new task: predicting future \emph{characteristic 3d poses} of human activities from short sequences of pose observations. 
We introduce a probabilistic approach to capturing the most likely modes in these characteristic poses, coupled with an autoregressive formulation for pose joint prediction to sample consistent 3d poses from a predicted joint distribution.
We trained and evaluated our approach on a new annotated dataset for characteristic 3d pose prediction, outperforming deterministic and multi-modal state-of-the-art approaches.
We believe that this opens up many possibilities towards goal-oriented 3d human pose forecasting and understanding anticipation of human movements.
\section*{Acknowledgements}
\label{sec:acknowledgements}

This project is funded by the Bavarian State Ministry of Science and the Arts and coordinated by the Bavarian Research Institute for Digital Transformation (bidt).

{\small
\bibliographystyle{ieee_fullname}
\bibliography{egbib}
}

\clearpage
\newpage
\begin{appendix}
\twocolumn[  
    \begin{@twocolumnfalse}
        \begin{center}
            {\Large Appendix}
            \vspace{0.5em}
         \end{center}
     \end{@twocolumnfalse}
]

In this appendix, we show additional qualitative results (Sec.~\ref{sec:additional-qualitative}), additional quantitative analysis (Sec.~\ref{sec:quantitative-results}), detail our network architecture specification (Sec.~\ref{sec:architecture}),  provide additional details regarding the dataset (Sec.~\ref{sec:dataset}) as well as our training setup (Sec.~\ref{sec:training}), and discuss potential negative societal impacts of our method (Sec.~\ref{sec:impacts}).

\section{Additional Qualitative Results.}
\label{sec:additional-qualitative}
\begin{figure*}
\begin{center}
	\includegraphics[width=\textwidth]{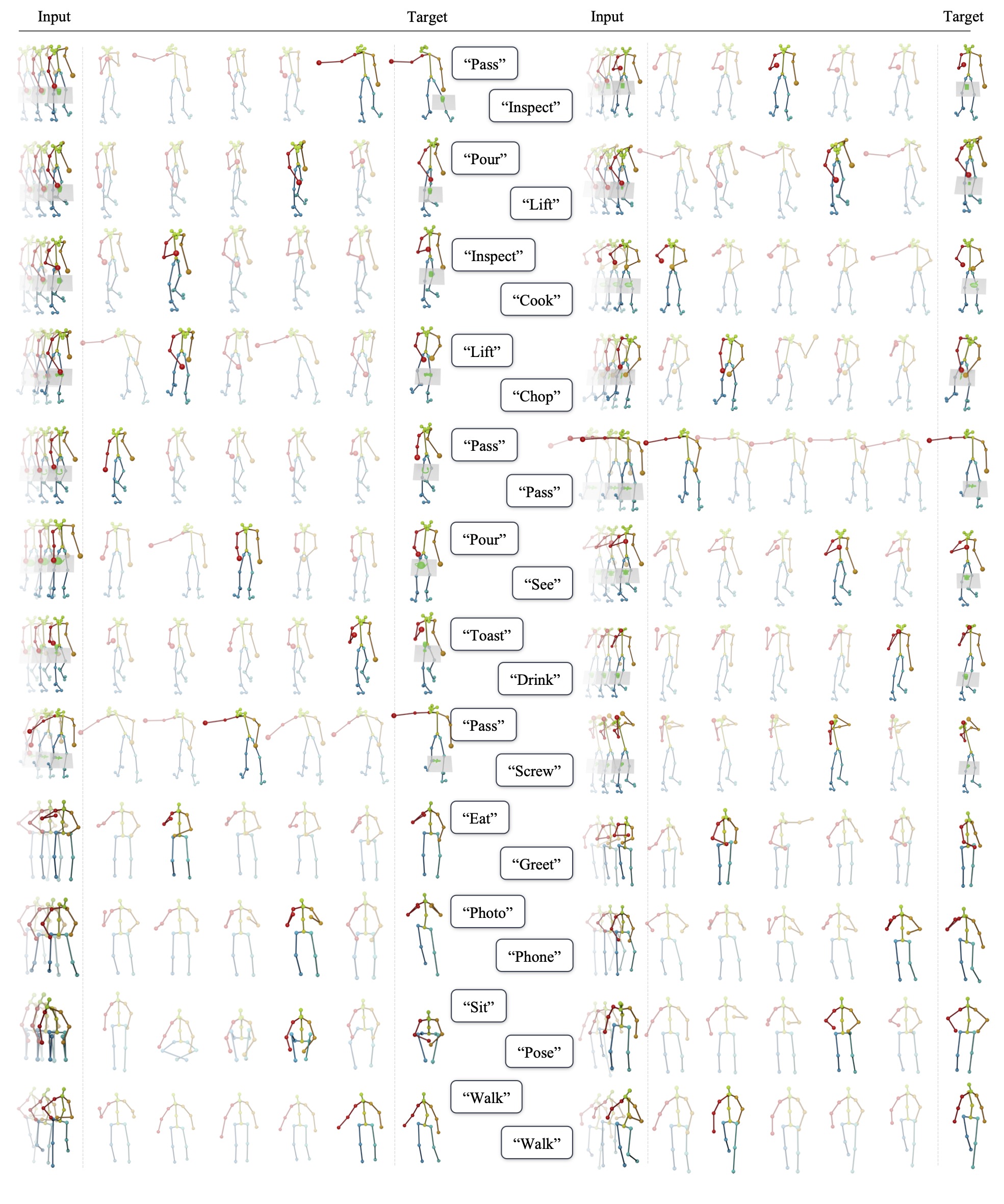}
\end{center}
\vspace{-0.6cm}
\caption{Additional qualitative results, showing the for each action sequence the inputs (left), our diverse set of predictions (middle) and the target action pose (right). Our final pose prediction is highlighted for each action sequence.}
\label{fig:additional-results}
\end{figure*}

We show additional qualitative results of our method in Fig. \ref{fig:additional-results}, which demonstrate the diversity of our characteristic pose predictions for a given input sequence. Our approach not only effectively models the multi-modal nature of characteristic poses, but also captures the final target action pose (highlighted pose prediction). 

In cases where the time between input sequence and target pose is longer, such as in ‘sit’ or ‘greet’, our approach produces a more diverse set of action poses, capturing the ambiguity in the future characteristic pose. When the input sequence is close to the target pose, our approach converges to a small set of probable poses (for example, in ‘drink’), reflecting the reduced ambiguity.

\section{Additional Quantitative Results.}
\label{sec:quantitative-results}

\begin{figure}[b]
\begin{center}
    \includegraphics[width=0.95\columnwidth]{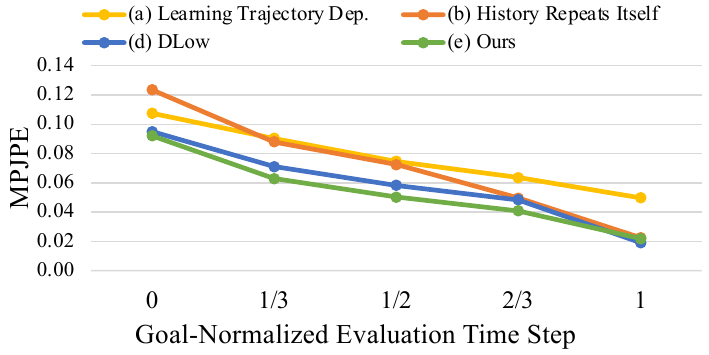}
\end{center}
\vspace{-0.6cm}
\caption{MPJPE comparison to baselines, evaluating with the input sequence at different points in time: from the start of the sequence (0) to $N$ frames before the target characteristic pose (1).}
\vspace{-0.5cm}
\label{fig:mpjpe_curves}
\end{figure}

\paragraph{MPJPE baseline comparison, by goal-normalized input time}
Fig. \ref{fig:mpjpe_curves} shows MPJPE for varying input sequence start times in comparison with state of the art, goal-normalized from the start of each sequence (0) to $N$ frames before the characteristic pose (1), with three steps inbetween. 

\paragraph{Autoregressive Joint Order.}
We determined the order of the joints for the autoregressive prediction empirically; most ambiguity occurred in active end-effectors (i.e. right and left hands), whereas the rest of the body tended to have lower variability. In Tab.~\ref{tab:ablation_grid_order}, we compare our original approach of (right hand, left hand, rest) with two alternatives:  (left hand, right hand, rest), and (full autoregressive from human kinematic chain following left/right hands).
Our method is robust to these orderings (though diversity of the rest of the body except hands decreases with autoregression through the kinematic chain).

\begin{table}[t]
\begin{center}
    \resizebox{0.9\linewidth}{!}{%
	\begin{tabular}{| l || c | c | c |}
		\hline
		 Order & \footnotesize MPJPE $\downarrow$ & \footnotesize Div. $\uparrow$ & \footnotesize IS $\uparrow$  \\ \hline\hline
        {\bf right hand $\rightarrow$ left hand $\rightarrow$ rest} & \textbf{0.054} & \textbf{0.105} & \textbf{4.15 {\footnotesize $\pm 0.9$}} \\ \hline\hline
        left hand $\rightarrow$ right hand $\rightarrow$ rest & 0.057 & 0.049 & 4.09 {\footnotesize $\pm 1.6$} \\ 
        following the kinematic chain & 0.058 & 0.018 & 4.02 {\footnotesize $\pm 0.9$}  \\ \hline
	\end{tabular}
	}
	\vspace{-0.1cm}
	\caption{Ablation analysis on autoregressive order on GRAB data.}
	\label{tab:ablation_grid_order}
\end{center}
\vspace{-0.66cm}
\end{table}

\paragraph{Grid Resolution and Offset Prediction.}
We show additional ablations on the effect of grid resolution and offset prediction in Tab~\ref{tab:ablation_grid_resolution} on GRAB data; A resolution of $16^3$ performs better than $8^3$ or $32^3$. Our offset prediction helps mitigate grid artifacts even at $32^3$.

\begin{table}[h]
\begin{center}
    \resizebox{0.9\linewidth}{!}{%
	\begin{tabular}{| c | c || c | c | c |}
		\hline
		Resolution & Offsets & \footnotesize MPJPE $\downarrow$ & \footnotesize Diversity $\uparrow$ & \footnotesize IS $\uparrow$ \\ \hline\hline
        $8^3$ & $\times$ & 0.242 & \textbf{0.189} & 1.40 {\footnotesize $\pm 0.3$} \\ 
        $8^3$ & $\checkmark$ & 0.092 & 0.068 & 1.71 {\footnotesize $\pm 0.1$}  \\ 
        $16^3$ & $\times$ & 0.127 & 0.081 & 1.51 {\footnotesize $\pm 0.1$} \\ \hline\hline
        $\boldsymbol{16^3}$ & $\checkmark$ & \textbf{0.054} & 0.105 & \textbf{4.15 {\footnotesize $\pm 0.9$}} \\ \hline\hline
        $32^3$ & $\times$ & 0.118 & 0.122 & 2.39 {\footnotesize $\pm 0.2$} \\ 
        $32^3$ & $\checkmark$ & 0.066 & 0.058 & 1.91 {\footnotesize $\pm 0.2$} \\ \hline
	\end{tabular}
	}
	\vspace{-0.1cm}
	\caption{Ablation analysis on heatmap grid size and offset prediction on GRAB data.}
	\label{tab:ablation_grid_resolution}
\end{center}
\vspace{-1cm}
\end{table}

\paragraph{Per-Bodypart MPJPE.}
In Tab.~\ref{tab:comparison-per-bodypart}, we show our final pose prediction performance in MPJPE, broken down per bodypart, as compared to sequential baselines.

\paragraph{Characteristic Pose Forecasting with Ground Truth Action Labels.}
In Tab.~\ref{tab:ablation-action-label}, we additionally evaluate our approach using ground truth action labels as input to provide additional contextual information. 

The ground truth action label is processed as an additional attention node alongside input and previously predicted joint locations.
This action label information reduces ambiguity in the possible set of output poses, resulting in reduced diversity, as is reflected in the diversity metric and inception score (as this directly considers diversity).

In our original action-agnostic scenario, our approach predicts plausible and diverse characteristic poses across all actions.
\begin{table}[H]
\begin{center}
	\resizebox{\linewidth}{!}{%
	\begin{tabular}{| l || c | c | c || c | c | c |}
	    \hline
	     & \multicolumn{3}{c||}{GRAB} & \multicolumn{3}{c|}{Human3.6M} \\
		\hline
		& MPJPE $\downarrow$ & Div. $\uparrow$ & IS $\uparrow$ & MPJPE $\downarrow$ & Div. $\uparrow$ & IS $\uparrow$ \\ \hline\hline
        $\times$ & 0.054 & 0.105 & 4.153 {\footnotesize $\pm 0.87$} & 0.092 & 0.189 & 3.139 {\footnotesize $\pm 0.32$} \\ \hline
        \checkmark & 0.051 & 0.026 & 1.085 {\footnotesize $\pm 0.02$} & 0.094 & 0.044 & 1.700 {\footnotesize $\pm 0.06$} \\ \hline
	\end{tabular}
	}
	\caption{Comparison of ours to an ablation with ground truth action labels as additional input.}
	\label{tab:ablation-action-label}
\end{center}
\vspace{-0.6cm}
\end{table}

\section{Architecture Details}
\label{sec:architecture}
\begin{figure*}[b]
\begin{center}
	\includegraphics[width=\linewidth]{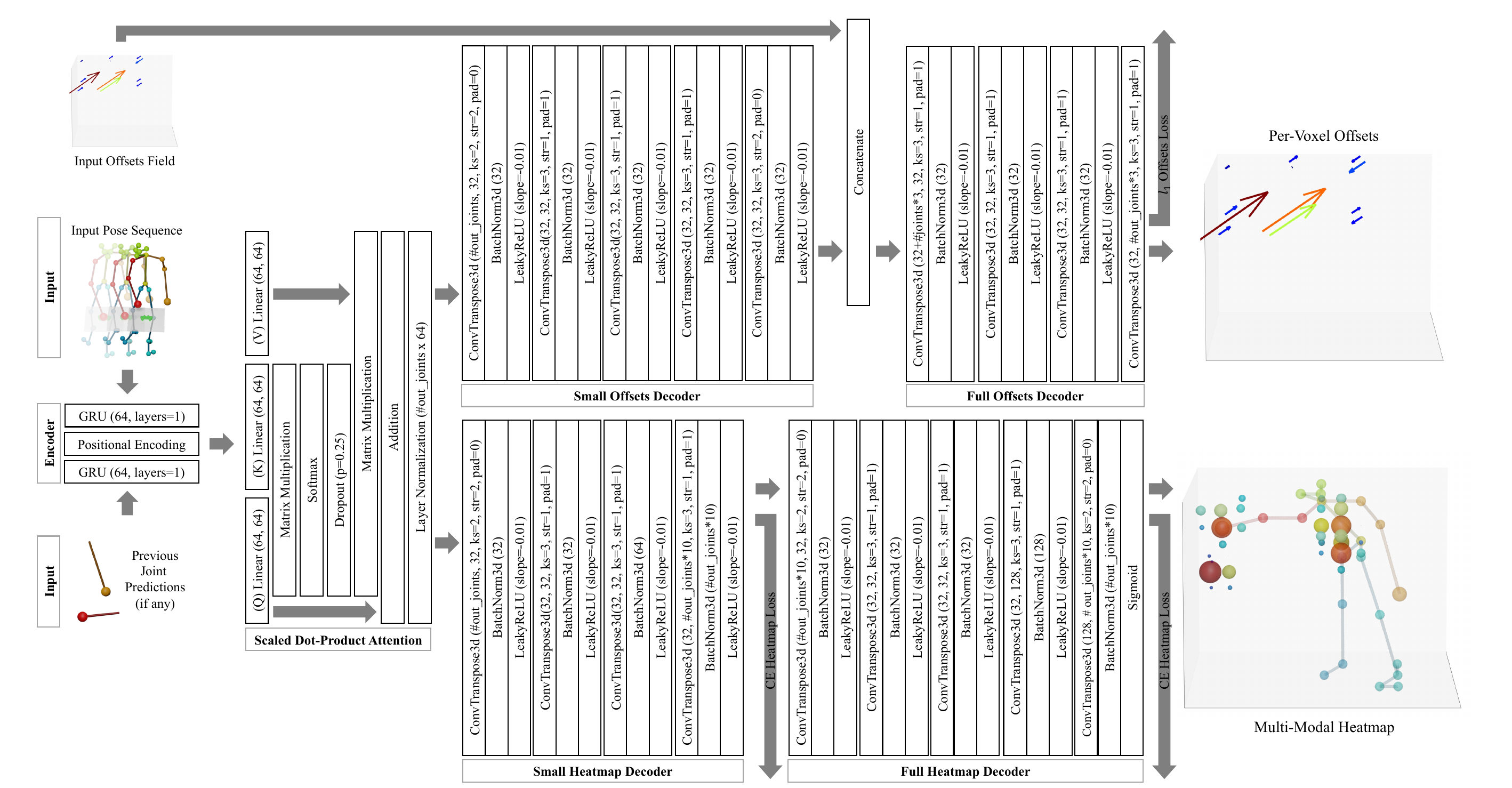}
\end{center}
\vspace{-0.3cm}
\caption{Our network architecture with details for encoder, scaled dot-product attention, as well as heatmap and offsets decoders.}
\label{fig:network}
\end{figure*}

Fig.~\ref{fig:network} details our network specification from input (left) to heatmap and offsets output (right). For each GRU layer, we provide the hidden dimension and number of layers in parentheses, for normalization layers the dimension to be normalized over, for dropout layers the dropout probability $p$, and for convolutions the number of input and output channels as well as kernel size (ks), stride (str), and padding (pad). We apply cross-entropy (CE) losses at a heatmap resolution of $8^3$ and at the final resolution of $16^3$; for the offsets prediction, we concatenate the offsets volume generated from the last input skeleton after 5 convolution blocks and supervise the final predictions with an $\ell_1$ loss.

We take as input $25$ joints in the case of GRAB and $17$ joints for Human3.6M (\#in\_joints). The number of output joints (\#out\_joints) depends on whether the right or left hand is being predicted (\#out\_joints=1) or the rest of the body (\#out\_joints=23 for GRAB, \#out\_joints=15 for Human3.6M). In all our experiments, we use 10 as the number of probability bins.

\section{Dataset}
\label{sec:dataset}

\begin{figure}[t]
\begin{center}
	\includegraphics[width=\linewidth]{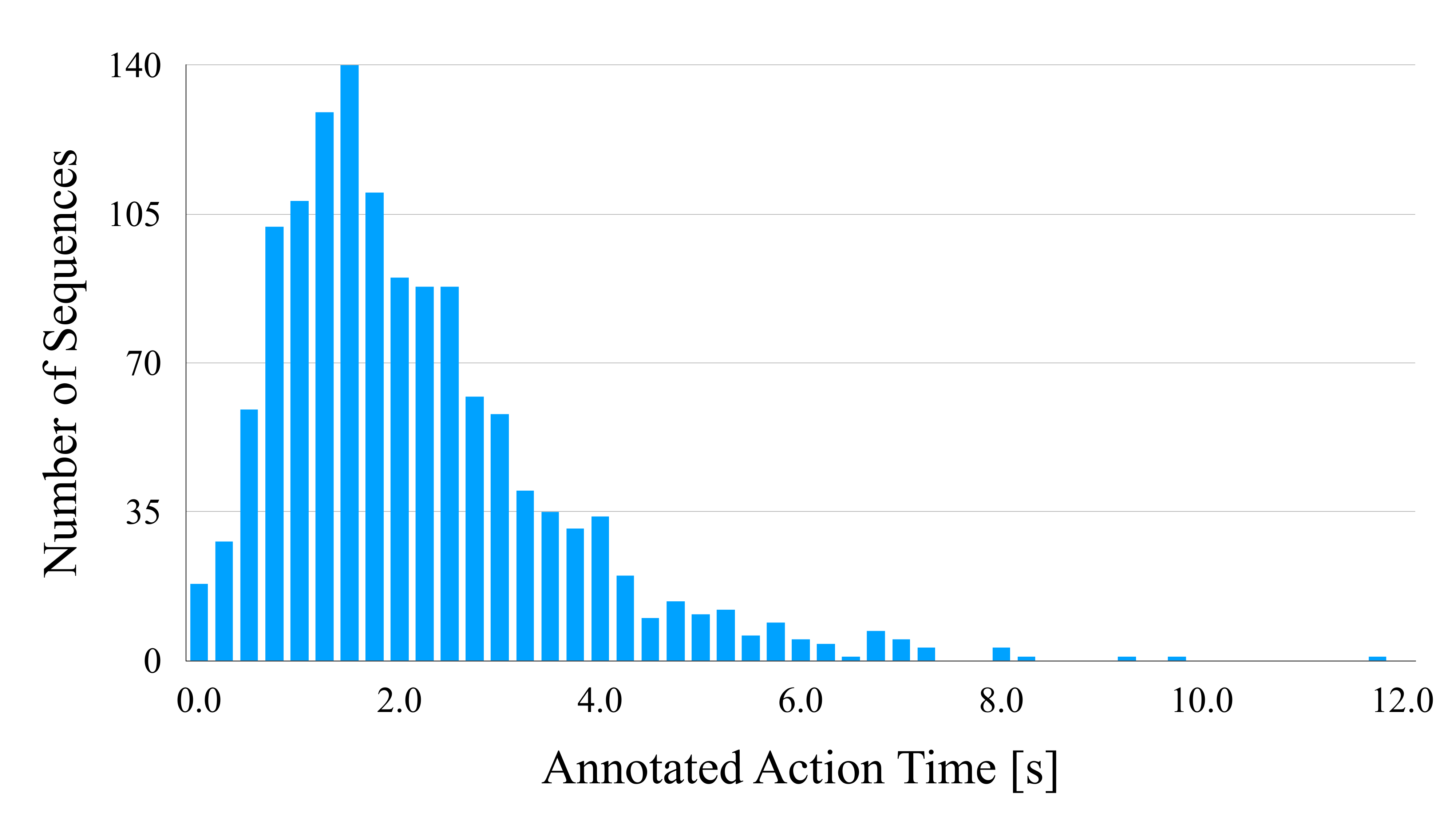}
\end{center}
\vspace{-0.6cm}
\caption{Times at which characteristic poses occur for GRAB.}
\vspace{-0.33cm}
\label{fig:action-annotation-timings-grab}
\end{figure}

\begin{figure}[t]
\begin{center}
	\includegraphics[width=\linewidth]{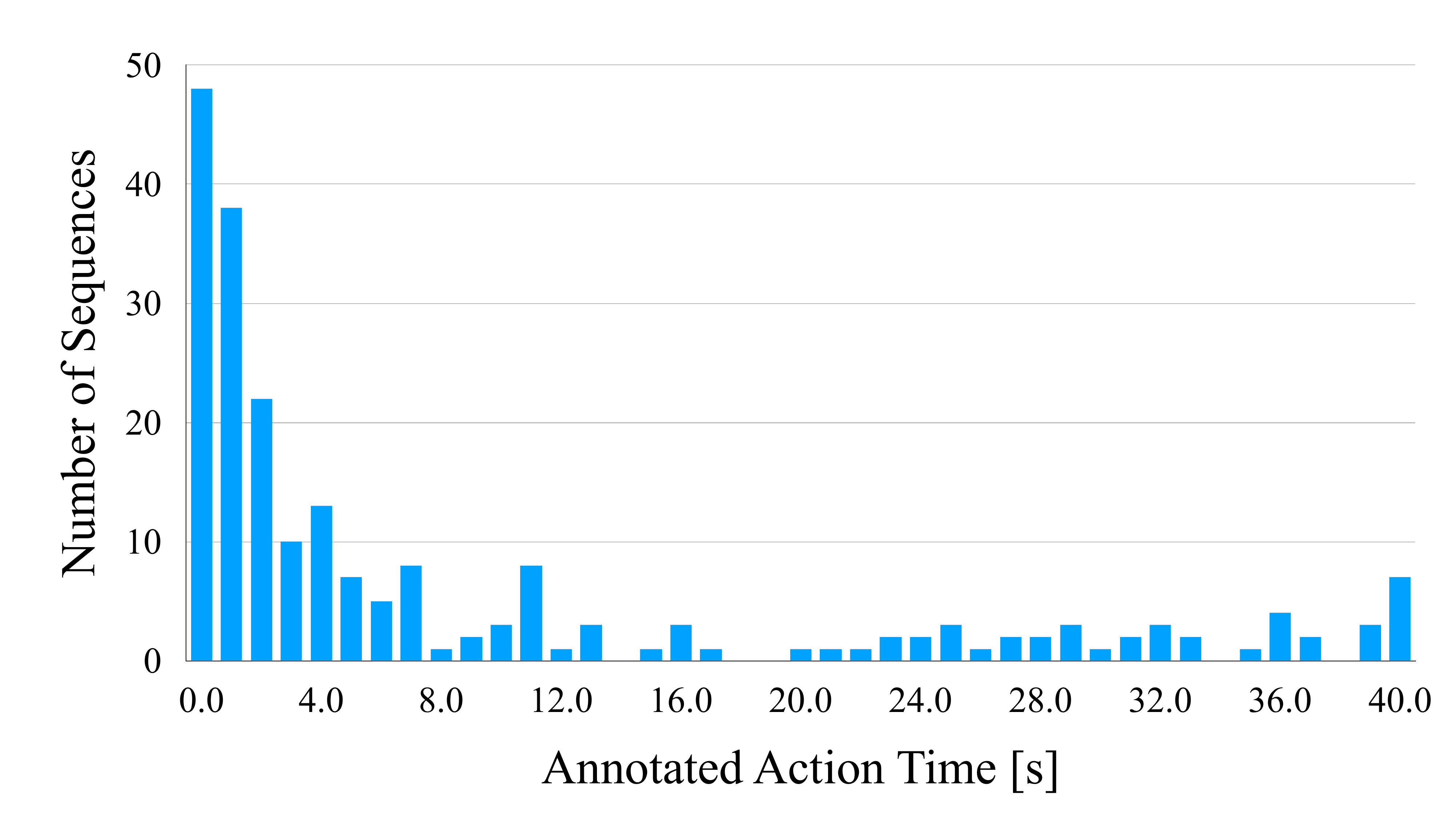}
\end{center}
\vspace{-0.6cm}
\caption{Times at which char. poses occur for Human3.6M.}
\vspace{-0.33cm}
\label{fig:action-annotation-timings-h36m}
\end{figure}

\begin{figure}[t]
\begin{center}
	\includegraphics[width=\linewidth]{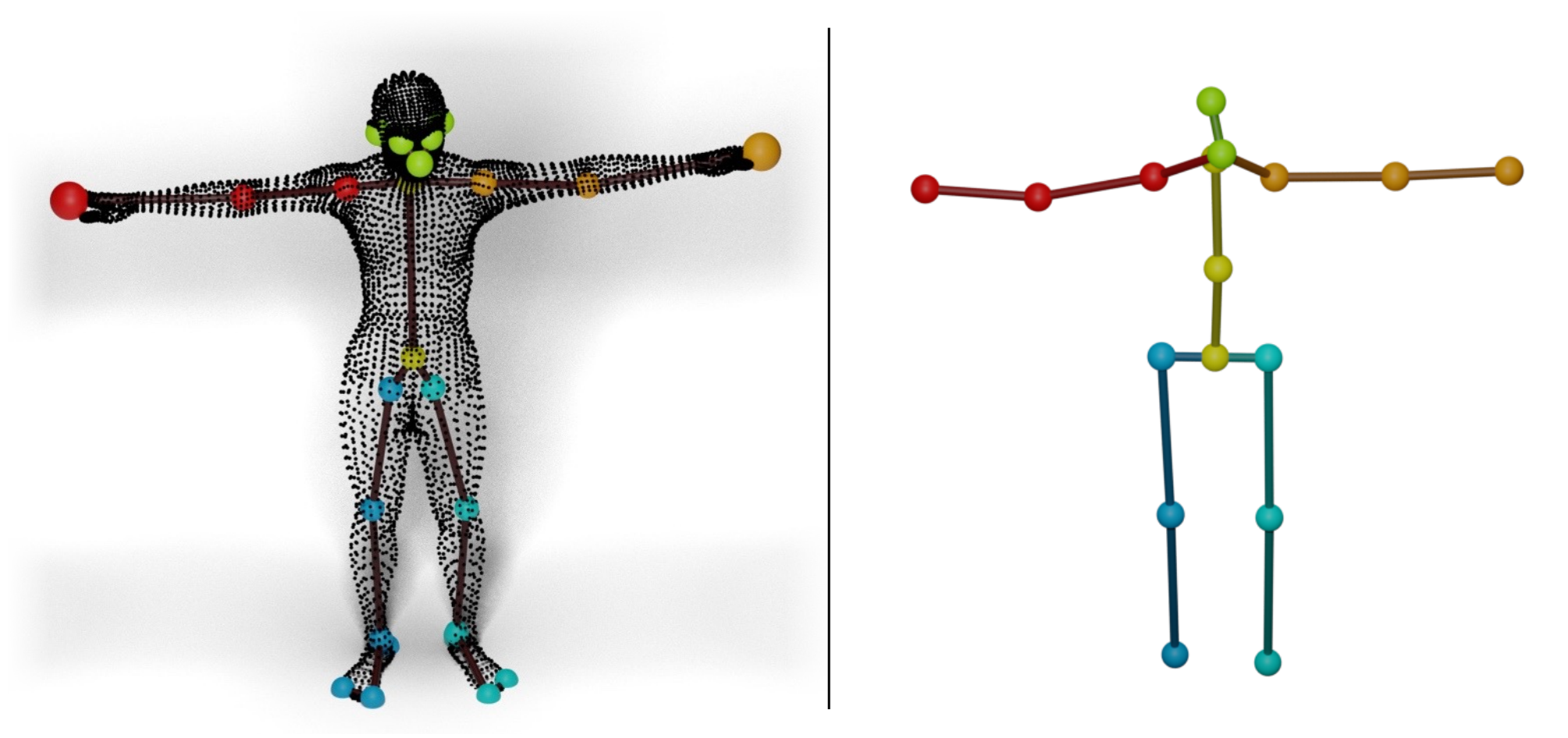}
\end{center}
\caption{GRAB~\cite{DBLP:conf/eccv/TaheriGBT20} body and our extracted skeleton joints overlaid (left); 17-joint skeleton based on Human3.6M~\cite{DBLP:journals/pami/IonescuPOS14} (right).}
\label{fig:body-combined}
\vspace{-0.5cm}
\end{figure}

\paragraph{GRAB Pose Layout.}
Since GRAB~\cite{DBLP:conf/eccv/TaheriGBT20} not only provides a human skeleton representation but full body shape parameters, we preprocess all pose sequences by first extracting relevant joints for our approach. For this, we chose the 3d OpenPose \cite{cao19OpenPose} layout as it describes the prevalent body joints and is widely used for representing 3d poses. Note that we do not apply the OpenPose method on 2d data; we only use their joint definitions in 3d.
We extract 25 body joints from the SMPL-X body given by the GRAB dataset~\cite{DBLP:conf/eccv/TaheriGBT20} using the correspondences shown in Tab.~\ref{tab:joint-correspondences-grab}. 
Additionally, we denote in Tab.~\ref{tab:joint-correspondences-grab} the correspondences of joints to body parts, for the body part analysis in Tab.~\ref{tab:comparison-per-bodypart}.
Fig.~\ref{fig:body-combined} (left) visualizes our joint selection, overlaying the body shape given in GRAB as a point cloud over the 25-joint skeleton.

\vspace{-1em}
\paragraph{Human3.6M Pose Layout.}
For all our experiments on Human3.6M~\cite{DBLP:journals/pami/IonescuPOS14}, we use 17 pose joints, visualized in Fig.~\ref{fig:body-combined} (right). Tab.~\ref{tab:joint-correspondences-h36m} describes the exact joints used as well as the correspondences of joints to body parts, as used in Tab.~\ref{tab:comparison-per-bodypart}.

\vspace{-1em}
\paragraph{Visualization Details.}
While our approach is agnostic to context or action, we visualize the context provided by GRAB~\cite{DBLP:conf/eccv/TaheriGBT20, Brahmbhatt_2019_CVPR} (of the table and object) and action label provided by both GRAB and Human3.6M to help contextualize the pose visualizations. 
The context and action labels are not taken into account by the network or the evaluation, meaning that our approach infers plausible human action poses while being agnostic towards action and context.

\begin{figure*}[t]
\vspace{-10cm}
\begin{center}
	\includegraphics[width=\linewidth]{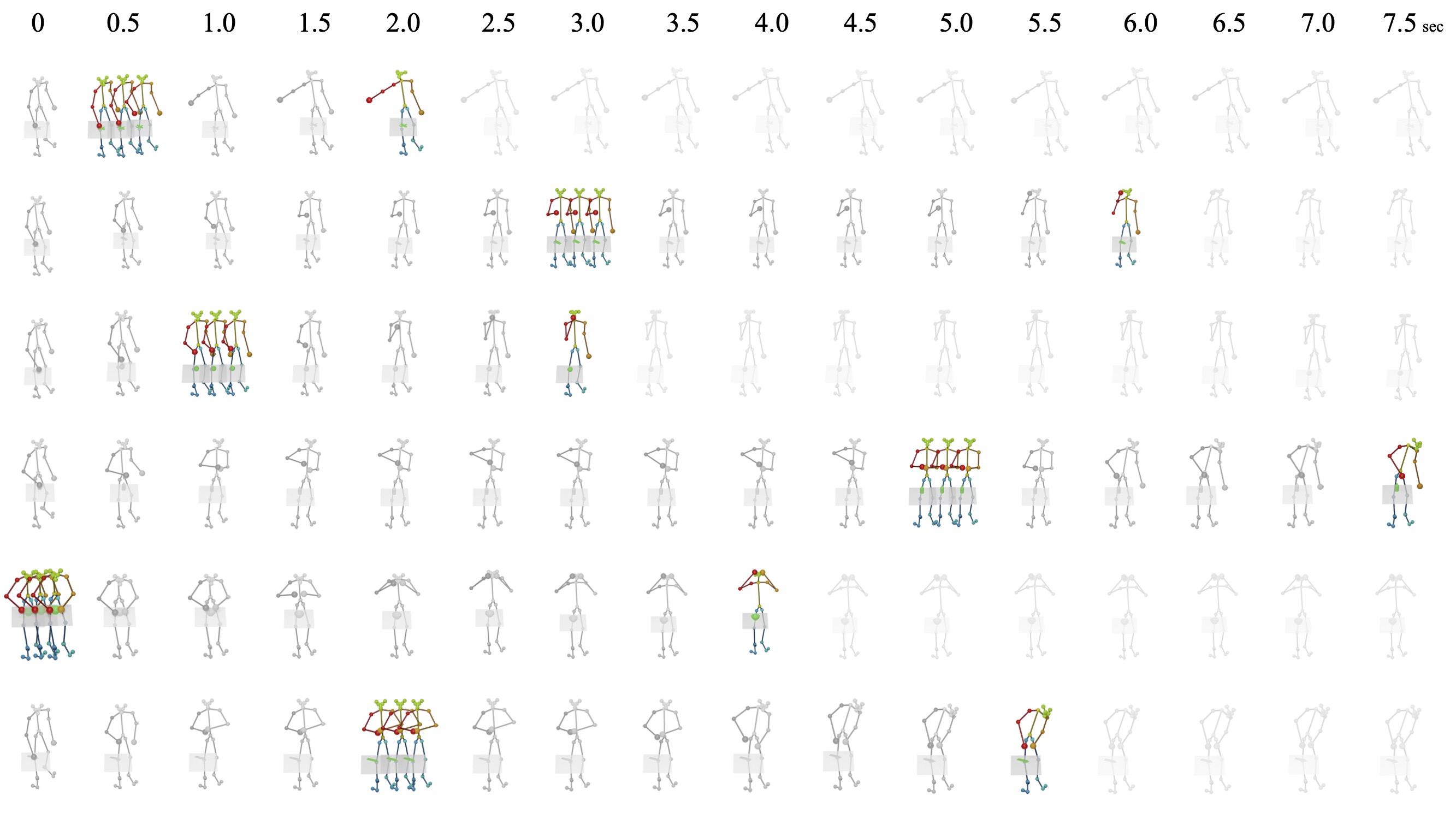}
\end{center}
\vspace{-0.6cm}
\caption{Sample input-target pairs (colored) for our characteristic 3d pose forecasting task, with temporal snapshots along the sequence (grayscale). Each snapshot is half a second apart. Depicted as input is the last frame of the respective input sequence.}
\vspace{-0.4cm}
\label{fig:filmstrips}
\end{figure*}

\vspace{-1em}
\paragraph{Additional Characteristic 3D Pose Details.}
We show additional characteristic 3d poses in their original sequences in Fig.~\ref{fig:filmstrips}, and note the strong time differences at which the characteristic poses occur.
Furthermore, Fig.~\ref{fig:action-annotation-timings-grab} and Fig.~\ref{fig:action-annotation-timings-h36m} show the times during the sequences at which the characteristic 3d poses are annotated for GRAB and Human3.6M; these characteristic poses are distributed across a wide range (0-12 seconds and 0-40 seconds, respectively) of time.

\section{Additional Training Details}
\label{sec:training}

\paragraph{Cross Entropy Loss.}
Since our approach learns to predict the probabilities of a Gaussian-smoothed target point during training, we observe a very large class imbalance between the no-probability bin (bin 0) and the rest of the bins. 
We thus weigh the classes in the cross entropy loss to account for the class imbalances, by  the inverse of their log-scaled occurrence, and a weight of $0.1$ for the no-probability bin.

\begin{table}[h]
\vspace{1.5em}
\begin{center}
\resizebox{0.8\linewidth}{!}{%
\begin{tabular}{| c || c | c || c | c |}
	\hline
	& \multicolumn{2}{c||}{Ours (17-Joint)} & \multicolumn{2}{c|}{Base (Human3.6M)} \\ \hline\hline
	& Idx & Label & Label & Idx \\ \hline\hline
	\multirow{3}{*}{\rotatebox[origin=c]{90}{R. Leg}} & 1 & R. Hip & R. Hip & 1 \\
	& 2 & R. Knee & R. Knee & 2 \\
	& 3 & R. Foot & R. Heel & 3 \\
	\hline
	\multirow{3}{*}{\rotatebox[origin=c]{90}{L. Leg}} & 4 & L. Hip & L. Hip & 6 \\
	& 5 & L. Knee & L. Knee & 7 \\
	& 6 & L. Foot & L. Heel & 8 \\
	\hline
	\multirow{3}{*}{\rotatebox[origin=c]{90}{R. Arm}} & 14 & R. Shoulder & R. Shoulder & 25 \\
	& 15 & R. Elbow & R. Elbow & 26 \\
	& 16 & R. Hand & R. Hand & 27 \\
	\hline
	\multirow{3}{*}{\rotatebox[origin=c]{90}{L. Arm}} & 11 & L. Shoulder & L. Shoulder & 17 \\
	& 12 & L. Elbow & L. Elbow & 18 \\
	& 13 & L. Hand & L. Hand & 19 \\
	\hline
	\multirow{2}{*}{\rotatebox[origin=c]{90}{Spine}} & 7 & Spine & Spine & 12 \\
	& 0 & Hip & Hip & 0 \\
	\multirow{3}{*}{\rotatebox[origin=c]{90}{Head}} & 9 & Nose & Nose & 14 \\
	& 10 & Head & Head & 15 \\
	& 8 & Thorax & Thorax & 13 \\
	\hline
\end{tabular}
}
	\caption{Joint Correspondences for Human3.6M}
	\label{tab:joint-correspondences-h36m}
\end{center}
\end{table}

\begin{table}[t]
\vspace{0.5em}
\begin{center}
\resizebox{0.85\linewidth}{!}{%
\begin{tabular}{| c || c | c || c | c |}
	\hline
	& \multicolumn{2}{|c||}{Ours (OpenPose \cite{cao19OpenPose})} & \multicolumn{2}{|c|}{Base (SMPL-X \cite{DBLP:conf/cvpr/PavlakosCGBOTB19})} \\ \hline
	& Idx & Label & Label & Idx \\ \hline\hline
	\multirow{3}{*}{\rotatebox[origin=c]{90}{R. Arm}} & 2 & Right Shoulder & Right Shoulder & 17 \\
	& 3 & Right Elbow & Right Elbow & 19 \\
	& 4 & Right Finger & Right Index 3 & 42 \\ \hline
	\multirow{3}{*}{\rotatebox[origin=c]{90}{L. Arm}} & 5 & Left Shoulder & Left Shoulder & 16 \\
	& 6 & Left Elbow & Left Elbow & 18 \\
	& 7 & Left Finger & Left Index 3 & 27 \\ \hline
	\multirow{6}{*}{\rotatebox[origin=c]{90}{Right Leg}} & 9 & Right Hip & Right Hip & 2 \\
	& 10 & Right Knee & Right Knee & 5 \\
	& 11 & Right Ankle & Right Ankle & 8 \\
	 & 22 & Right Big Toe & Right Big Toe & 63 \\
	& 23 & Right Small Toe & Right Small Toe & 64 \\
	& 24 & Right Heel & Right Heel & 65 \\ \hline
	\multirow{6}{*}{\rotatebox[origin=c]{90}{Left Leg}} & 12 & Left Hip & Left Hip & 1 \\
	& 13 & Left Knee & Left Knee & 4 \\
	& 14 & Left Ankle & Left Ankle & 7 \\
	& 19 & Left Big Toe & Left Big Toe & 60 \\
	& 20 & Left Small Toe & Left Small Toe & 61 \\
	& 21 & Left Heel & Left Heel & 62 \\ \hline
    \multirow{6}{*}{\rotatebox[origin=c]{90}{Head}} & 0 & Nose & Nose & 55 \\
	& 1 & Neck & Neck & 12 \\
	& 15 & Right Eye & Right Eye & 24 \\
	& 16 & Left Eye & Left Eye & 23 \\
	& 17 & Right Ear & Right Ear & 58 \\
	& 18 & Left Ear & Left Ear & 59 \\ \hline
	& 8 & Mid-Hip & Pelvis & 0 \\
	\hline
\end{tabular}
}
	\caption{Joint Correspondences for GRAB}
	\label{tab:joint-correspondences-grab}
\end{center}
\end{table}

\afterpage{\clearpage}

\clearpage

\begin{table*}[t]
\begin{center}
	\small
	\resizebox{\linewidth}{!}{%
	\begin{tabular}{| l || c | c | c | c | c | c | c | c | c | c | c | c |}
	    \hline
	    & \multicolumn{6}{|c|}{GRAB} & \multicolumn{6}{c|}{H3.6M} \\ \hline
		Method & R. Arm $\downarrow$ & L. Arm $\downarrow$ & R. Leg $\downarrow$ & L. Leg $\downarrow$ & Spine $\downarrow$ & Head $\downarrow$ & R. Arm $\downarrow$ & L. Arm $\downarrow$ & R. Leg $\downarrow$ & L. Leg $\downarrow$ & Spine $\downarrow$ & Head $\downarrow$ \\ \hline\hline
        L. T. D.~\cite{DBLP:conf/iccv/MaoLSL19} & 0.165 & 0.115 & 0.058 & 0.057 & 0.028 & 0.085 & 0.225 & 0.225 & 0.135 & 0.146 & 0.108 & 0.123 \\ \hline
        H. R. I.~\cite{DBLP:conf/eccv/MaoLS20} & 0.160 & 0.113 & 0.056 & 0.055 & 0.026 & 0.079 & 0.199 & 0.191 & \textbf{0.079} & 0.088 & 0.040 & 0.089 \\ \hline
        DLow \cite{yuan2020dlow} & 0.146 & 0.109 & 0.052 & 0.050 & 0.024 & 0.068 & 0.174 & 0.169 & 0.108 & 0.112 & 0.044 & 0.096 \\ \hline
        \textbf{Ours} & \textbf{0.105} & \textbf{0.084} & \textbf{0.045} & \textbf{0.045} & \textbf{0.020} & \textbf{0.057} & \textbf{0.147} & \textbf{0.122} & 0.091 & \textbf{0.085} & \textbf{0.033} & \textbf{0.066} \\ \hline
	\end{tabular}
	}
	\vspace{-0.1cm}
	\caption{Characteristic 3d pose prediction performance comparison to baselines, broken down by body part MPJPE.}
	\vspace{-0.5cm}
	\label{tab:comparison-per-bodypart}
\end{center}
\end{table*}

\paragraph{State-of-the-art comparisons.}
We use the official code with default settings of the methods we compare to (\cite{DBLP:conf/iccv/MaoLSL19}, \cite{DBLP:conf/eccv/MaoLS20}, and \cite{yuan2020dlow}). We train all methods from scratch on our characteristic 3d pose dataset, setting the number of input frames to 10 and the number of output frames to 100. From the predicted sequence, we evaluate the pose at a timestep predicted by the baselines themselves as characteristic pose and compare it to the target. This scenario is the closest to our approach, as predicting characteristic 3d poses involves which pose is the characteristic pose.

Therefore, we  modified each baseline with a small prediction head to predict the characteristic pose frame within all 100 frames of the predicted sequence. In all cases, we supervise this prediction as a classification problem with a cross entropy loss and train the additional head together with the rest of the model.

For DLow \cite{yuan2020dlow}, we add one linear layer to the final feature output of each of the 100 steps, followed by a ReLU, reducing each step's output dimension to 10. Then, one additional linear layer summarizes the combined output of all steps ($100*10$) down to a vector of size 100.

In the case of History Repeats Itself \cite{DBLP:conf/eccv/MaoLS20}, we add a classification head consisting of one linear layer, a 1d batch norm, a ReLU, and one additional linear layer to the output of their last Graph Convolution Block (GCN). While the first linear layer keeps the original dimensionality of $100$, the second linear layer reduces the dimension from $\text{\#graph\_nodes}*100$ down to $100$.

Finally, for Learning Trajectory Dependencies \cite{DBLP:conf/iccv/MaoLSL19}, we apply the same architecture and add a linear layer, a 1d batch norm, a ReLU, and a second linear layer after the final GCN. Here, we first reduce the per-node feature dimension from 256 to 100 and combine the features of all nodes with the second linear layer, going from $\text{\#graph\_nodes}*100$ down to $100$.

In the main paper, we additionally evaluated against these baseline approaches when given ground-truth time steps instead; in this scenario, our predictions also outperform the baselines given ground truth times for characteristic poses.

To evaluate the diversity and quality of multi-modal outputs, 10 samples are taken from a probabilistic method for each input sequence, and we report diversity in terms of MPJPE between samples as well as the Inception Score, following \cite{aliakbarian2020stochastic}.

\section{Potential Negative Societal Impacts}
\label{sec:impacts}

As we aim to study human pose behavior, we must take care to ensure that datasets used represent notable diversity in those represented.
Our approach currently operates on skeleton abstractions that do not characterize finer-scale appearance differences; in possible future studies that may aim to characterize fine-scale interactions, diversity in body shape representations which must be taken into account for data collection and analysis.

In particular, in our scenario of forecasting probable future human behavior, we must also ensure that this possibility cannot be easily used for generating fraudulent motion video of a person. Such usage is currently severely limited in our proposed approach, as it does not target individual people, and does not model photo-realistic characteristics of people. 

Another concern might arise with the possibility of surveillance, in the context of predicting specific actions from only a short and possibly ambiguous observation of a person. The types of actions are currently limited by the training data to everyday activities such as eating or walking. With modified datasets, the prediction of various specific action sub-categories might be possible (e.g., forecasting possible malicious actions). While simpler methods may be more suitable for this kind of task, here we look to efforts in data transparency; we will provide our annotations and various statistics to characterize the everyday activities in our considered data.

Another axis to consider is that of environmental impact, in the cost of training deep neural networks. Our training time is relatively short with only a few hours until convergence and a moderately sized neural network. 
Additionally, adversarial attacks are a possibility to disrupt future predictions, but do not induce security concerns for our approach directly.

\clearpage

\end{appendix}

\end{document}